\documentclass[conference]{IEEEtran}
\IEEEoverridecommandlockouts
\usepackage{cite}
\usepackage{lmodern}
\usepackage[utf8]{inputenc}
\usepackage{amsmath,amssymb,amsfonts}
\usepackage{algorithmic}
\usepackage{graphicx}
\usepackage{textcomp}
\usepackage{xcolor}
\usepackage{multirow}
\usepackage{hyperref}
\def\BibTeX{{\rm B\kern-.05em{\sc i\kern-.025em b}\kern-.08em
		T\kern-.1667em\lower.7ex\hbox{E}\kern-.125emX}}

\begin{document}

	
	\title{Inference of Recyclable Objects with Convolutional Neural Networks
		
	\vspace{0.75cm}

	\small \textit{This document is a pre-print version in English language. \\ The original version will be published in Spanish at the \href{https://revistas.utp.ac.pa/index.php/ric}{RIC Journal } (P-ISSN 2412-0464/E-ISSN 2413-6786) of the Technological University of Panama.}

	
}
	
	\makeatletter
	\newcommand{\linebreakand}{%
	\end{@IEEEauthorhalign}
	\hfill\mbox{}\par
	\mbox{}\hfill\begin{@IEEEauthorhalign}
	}
	\makeatother
	
	\author{
		
		\IEEEauthorblockN{1\textsuperscript{st} Jaime Caballero}
		\IEEEauthorblockA{\textit{Faculty of Electrical Engineering} \\
			\textit{Technological University of Panama}\\
			David City, Panama \\
			\href{mailto:jaime.caballero1@utp.ac.pa}{jaime.caballero1@utp.ac.pa}}
		
		\and
		
		\IEEEauthorblockN{2\textsuperscript{nd} Francisco Vergara}
		\IEEEauthorblockA{\textit{Faculty of Electrical Engineering} \\
			\textit{Technological University of Panama}\\
			David City, Panama \\
			\href{mailto:francisco.vergara@utp.ac.pa}{francisco.vergara@utp.ac.pa}}
		
		\and
		
		\IEEEauthorblockN{3\textsuperscript{rd} Randal Miranda}
		\IEEEauthorblockA{\textit{Faculty of Electrical Engineering} \\
			\textit{Technological University of Panama}\\
			David City, Panama \\
			\href{mailto:randal.miranda@utp.ac.pa}{randal.miranda@utp.ac.pa}}
			
		\linebreakand
			

		\IEEEauthorblockN{Jos\'{e} Serrac\'{i}n, Ph.D.}
		\IEEEauthorblockA{{\footnotesize \textit{-Corresponding Author-}} \\ \textit{Faculty of Electrical Engineering} \\
			\textit{Technological University of Panama}\\
			David City, Panama \\
			\href{mailto:jose.serracin@utp.ac.pa}{jose.serracin@utp.ac.pa}}
	}
	
	\maketitle
	
	\begin{abstract}
		
		Population growth in the last decades has resulted in the production of about 2.01 billion tons of municipal waste per year. The current waste management systems are not capable of providing adequate solutions for the disposal and use of these wastes.  Recycling and reuse have proven to be a solution to the problem, but large-scale waste segregation is a tedious task and on a small scale it depends on public awareness. This research used convolutional neural networks and computer vision to develop a tool for the automation of solid waste sorting. The Fotini10k dataset was constructed, which has more than 10,000 images divided into the categories of ``plastic bottles'', ``aluminum cans'' and ``paper and cardboard''. ResNet50, MobileNetV1 and MobileNetV2 were retrained with ImageNet weights on the Fotini10k dataset. As a result, top-1 accuracy of 99\% was obtained in the test dataset with all three networks. To explore the possible use of these networks in mobile applications, the three nets were quantized in float16 weights. By doing so, it was possible to obtain inference times twice as low for Raspberry Pi and three times as low for computer processing units. It was also possible to reduce the size of the networks by half. When quantizing the top-1 accuracy of 99\% was maintained with all three networks. When quantizing MobileNetV2 to int-8, it obtained a top-1 accuracy of 97\%. \\
		
	\end{abstract}
	
	
	\section{Introduction}

		Since the 20th century, society has been in an endless cycle of using and throwing away. According to a study carried out by the World Bank in 2017, currently around 2.01 billion tons of municipal solid waste (MSW) are generated per year, of which more than 33\% are not treated in an adequate manner. It is also expected that by 2030 3.40 billion tons of MSW will be generated per year \cite{kaza2018waste}. This situation produces a global problem in terms of environmental pollution, social inclusion and economic sustainability. These problems affect both developed countries as well as developing countries, but their effects are much more pronounced on the latter. Due to the rapid growth of the urban centers in developing countries and poor waste management (WM), an uncontrolled disposition of waste in open spaces has been developed. These are popularly known as landfills, which may seem like a viable solution, but numerous studies such as \cite{ferronato2019waste} have shown that these produce problems such as: visual impact, air pollution, odors, greenhouse gases, disease vectors, surface and underground water contamination. \\
		
		This research proposes a tool to increase the efficiency of waste segregation systems in all scales and thus simplify the task of recycling. The proposed tool is the use of artificial intelligence (AI). Specifically, the use of convolutional neural networks (CNN) to develop and implement techniques of classification and detection of recyclable objects in images to automate their sorting. \\
		
		The use of CNN for classification tasks in computer vision has been practically the standard since 2012, after the seminal network AlexNet \cite{alexnet} popularized its use by winning Imagenet Challenge: ILSVRC 2012 \cite{imagenet} using a novel architecture based on deep convolutional layers. In the following years more complex architectures were developed - VGG \cite{vgg}, GoogLeNet \cite{googlenet}, ResNet \cite{resnet} - with the aim of obtaining better classification accuracy. \\
		
		The use of CNN for waste classification has already been proposed in other investigations. Awe et al. \cite{stanford} implemented the Faster R-CNN object detection model to detect waste. He obtained a mAP of 0.683. Gyawali et al. \cite{nepal} used CNNs, specifically ResNet50, VGG16 and ResNet18. The best result was obtained with ResNet18, with a top-1 precision of 87.8\%. \\
		
		The objective of this work is to introduce the Fotini10k dataset. Showcase its robustness based on its large quantity of images representative of real scenes. Another goal is to validate the performance of CNNs trained with the Fotini10k dataset and comparing performance against previous works on waste classification. And thus, provide affordable/innovative tools for automatic waste sorting that can be used on embedded devices. \\
		
		The relevance of this research is to demonstrate that the results are a firm starting point for improving waste segregation systems. CNNs could be used both in mixed recycle facility (MRF) and in day-to-day situations, through the automation of the waste sorting task. It should be noted that sorting waste is a cognitively simple task, but it becomes highly tedious on a large scale and on a small scale the complexity will depend on the culture or desire of the people. \\
		
		Section \ref{m_m} will address the premises taken for the elaboration of the dataset, in addition to introducing the architectures and methods used to train the implemented CNNs. In Section \ref{r_d} the results obtained with the different networks and their variants will be presented, as well as comparisons of precision and speed of inference. Section \ref{concl} will present the conclusions, value, relevance and future research work.

	\section{Materials and Methods}
	\label{m_m}
	
	The next subsection explains how the neural networks work and how they are used to make inference in images.  

	\subsection{CNNs for Computer Vision}
	
	Deep Learning (DL) is a set of techniques and algorithms that model high-level representations. The model used is a parametric function, made up of various types of operations that in turn require parameter setting or weights in an iterative process known as learning or training. Mathematically this learning is achieved by defining a loss function (it quantifies the error) to the parametric function and it optimizes for loss reduction. In DL the parametric function is an artificial neural network (ANN). \\
	
	One type of ANN are CNNs. These are used when the data to be processed can be represented in the form of matrix. In computer vision the processed images are digitally represented as an array of pixels with \verb|shape = [height, width, (light_channels)]| CNNs for computer vision perform the convolution operation between an image and a filter to generate a feature map \cite{rosebrock_dl4cv}. This operation of convolution in conjunction with activation functions, such as ReLu, and other operations such as Pooling, DropOut, Batch Normalization among others, are what is called layers and are components that define the architecture of a CNN. The CNN can classify, detect and infer semantic information of the content of an image. To achieve this, the CNN needs to pre-train with a great number of images. The group of these images are called the training dataset. Each iteration of training with all the images from the training dataset is called \verb|epoch|. During the training the loss and accuracy are monitored in the training dataset and generally also in a dataset of validation. The validation dataset must not contain images from the training dataset. The objective of monitoring is to indicate that the loss has converged and the accuracy increased.  The standard loss function in CNN is the function of loss crossed entropy. The most common optimizers for training are: stochastic gradient descent (SDG), Adam \cite{adam}, RMSprop \cite{rmsprop}. For the performance comparison of the different architectures, datasets like Imagenet  \cite{imagenet}, COCO \cite{coco} are commonly used. \\
	
	\subsubsection{Image Classification and Object Detection}
	
	The most common tasks performed by CNNs in computer vision are image classification (IC) and object detection (OD). IC consists in a model that assigns a label to an image from a discrete number of predefined categories, as seen in Fig.~\ref{fig1} on the left. OD is similar to IC, but with the additional step that it also provides the location of the object in the image, as illustrated in Fig.~\ref{fig1} to the right. To implement OD, traditional computer vision techniques such as Haar cascades, HOG + SVM, etc. are used to extract the location of the object in the image and then a CNN from IC assigns the label. Alternatively CNNs models such as Single Shot Detectors (SDD) \cite{ssd} do this operation in a single forward pass.
	
	\begin{figure}[htbp]
		\centerline{\includegraphics[width=\linewidth]{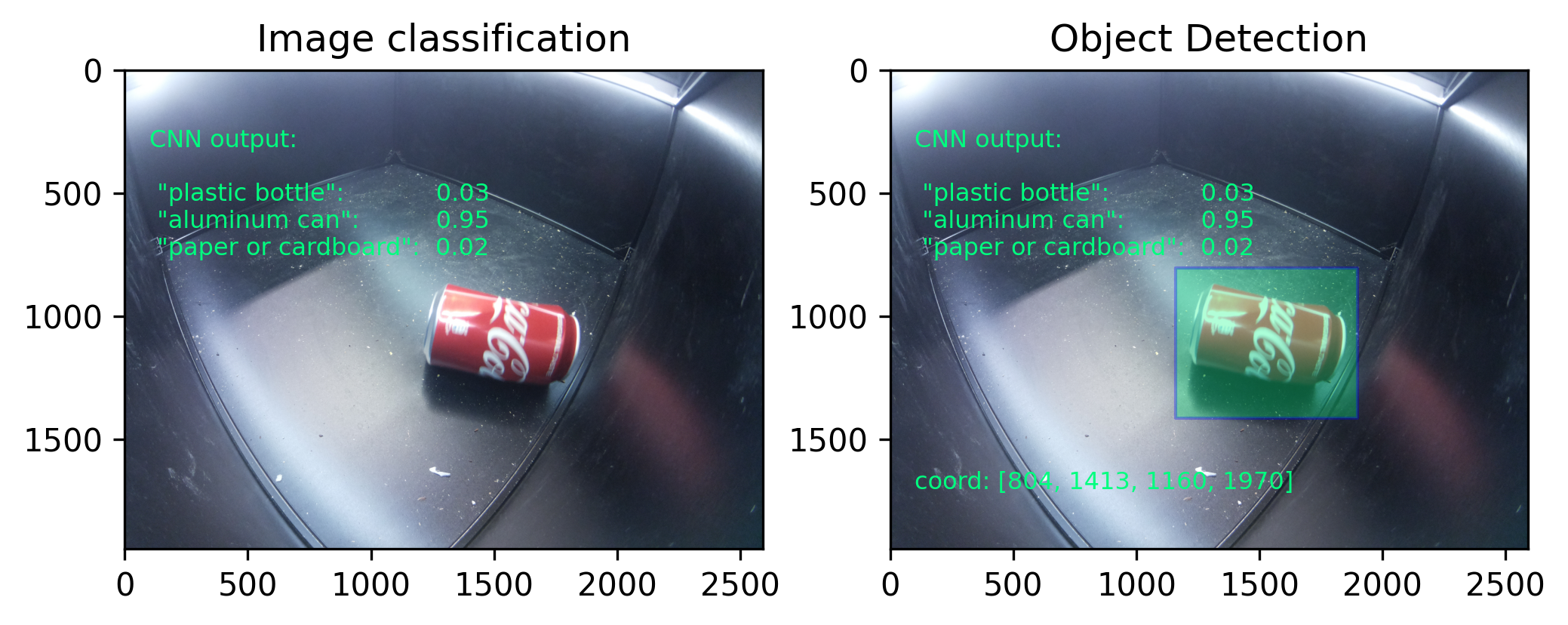}}
		\caption{On the left in green text is the results of applying IC to this image. On the right are the results obtained by applying OD.}
		\label{fig1}
	\end{figure}

	\subsubsection{Transfer Learning}
	
	Networks trained in the ImageNet dataset can classify the content of an image among 1000 common categories like dogs, apples, cars; which is not optimal for a specific application. For the implementation that this work intends is favorable that the trained network is very accurate among a few categories of waste. In this situation, it is common to use transfer learning, specifically fine tuning. This consists in modifying and retraining just the last layers of a CNN which was previously trained on a dataset with similar categories to the ones intended to classify. The use of this technique results in fewer less images and epochs to obtain high performance. \\
	
	All CNNs used in this investigation are previously trained in the ImageNet dataset and the last layers are retrained with the Fotini10k dataset. The specific architectures retrained and their respective modifications in the last layers are mentioned in Section \ref{resnet} and Section \ref{mobilenet_ar}. \\

	\subsubsection{Quantization}
	
	A network like ResNet50 has 18M parameters and occupies 92 MB of memory. The size of this file is small in the context of common computers, but this work proposes the use of CNNs on mobile or embedded devices. These have very little memory, in the order of a few tens of MB, and limited computational resources. Most ML frameworks store the weights of a network as \verb|32-bit| floating point constants (float32, FP32). A technique called quantization \cite{quan} allows you to convert the weights \verb|32-bit| to \verb|16-bit| (float16, FP16, f16). They can even be converted to \verb|8-bit| unsigned integers (UI8, i8), to be compatible with \verb|8-bit| architecture on microcontrollers or tensor processing units (TPU). Using this technique allows networks to occupy \verb|2x| to \verb|4x| less space in memory and increase the inference speed from \verb|2x| to \verb|4x|. The disadvantage is that the overall precision decreases and that not all network architectures support this conversion. \\
	
	Precision and inference time (time it takes the network to classify a certain amount of images) performance of the CNNs was compared. The comparisons were made among the CNNs implemented in their standard \verb|32-bit|  form and quantized to \verb|16-bit| weights and \verb|8-bit| weights.
	
	\subsection{Fotini10k Dataset}
	
	TrashNet \cite{stanford} and Gywali et al. \cite{nepal} used datasets of a limited number of images (1000)  per category. The images used by them were taken from the web and images of objects in different backgrounds. In contrast, our models are intended to be used in enclosures where you can control, to some degree, the visual conditions. In this way high classification accuracy is easier to obtain. \\
	
	For this reason, a robot was built to capture images of recyclable objects in a visually neutral box-like enclosure. Different light intensities were used while taking the images of  the Fotini10k dataset. In this way the enclosure resembles conditions that can be replicated in a recycling plant or smart trash bin. \\
	
	Previous works focused on classifying the categories: \verb|plastic, metal, organic, etc|. These categories depend on the composition of the material and not necessarily from its visual aspect. This work proposes to classify between the categories: \verb|'plastic bottles', 'aluminum cans'| and \verb|'paper and cardboard'|. The reason is that these categories do depend on the visual aspect and also make up most of the demand of recyclable objects. \\
	
	Given these considerations, the Fotini10k dataset was developed with around 3000 images per category and 10391 images in total. The resolution of the images is 2592 x 1944 and each image is annotated with the coordinates of the location of the object. Fig.~\ref{fig2} shows examples of the images in their standard variant at the left and cropped to the right. \\
	
	\begin{figure}[htbp]
		\centerline{\includegraphics[width=\linewidth]{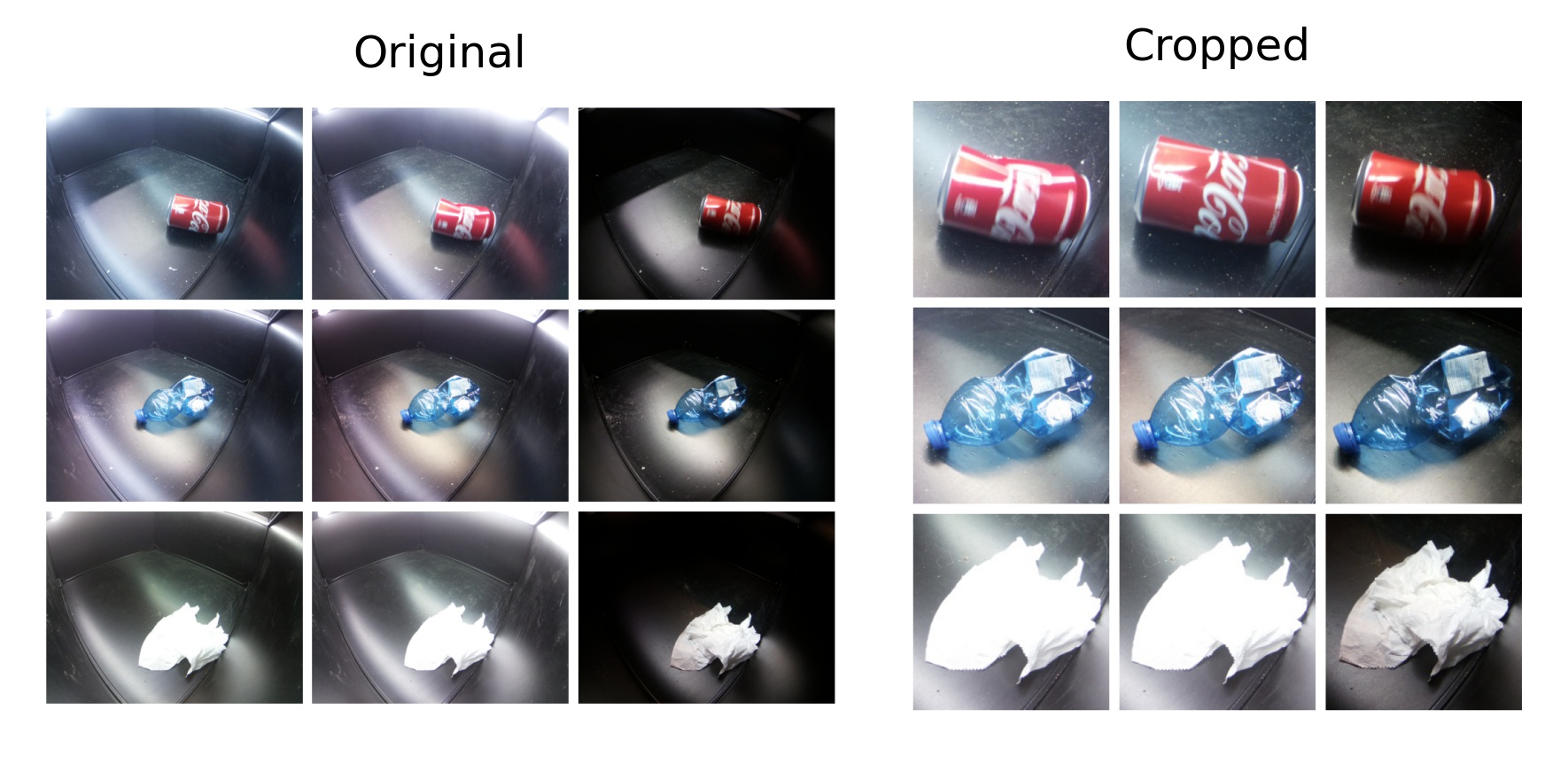}}
			\caption{Examples of images from the Fotini10k dataset with different levels of brightness are shown on the left. On the right the images are in its cropped version.  The coordinates show where the object is located within the image.}
		\label{fig2}
	\end{figure}

	During the training and validation of the CNNs, 60\% (6391) of the images were used for the train set, 20\% (2000) for the validation set and the remaining 20\% (2000) for the test set. Every image belongs only to one of the three sets. \\
	
	In experiment \#1 in Section \ref{exp1} the precision of a CNN is compared when training it with the images in original form (without taking aspect ratio into account) vs cropped (taking aspect ratio into account). \\
	
	\pagebreak
	
	\subsection{ResNet}
	\label{resnet}
	
	ResNet implements a microarchitecture residuals map \cite{resnet}, for this work we used ResNet50. In this way a direct comparison can be made to the performance of this architecture with our dataset to ResNet based models used in previous investigations. The modifications that were made to the last layers of ResNet50 are shown in Fig.~\ref{fig3}. 
	
	\begin{figure}[htbp]
		\centerline{\includegraphics[width=\linewidth]{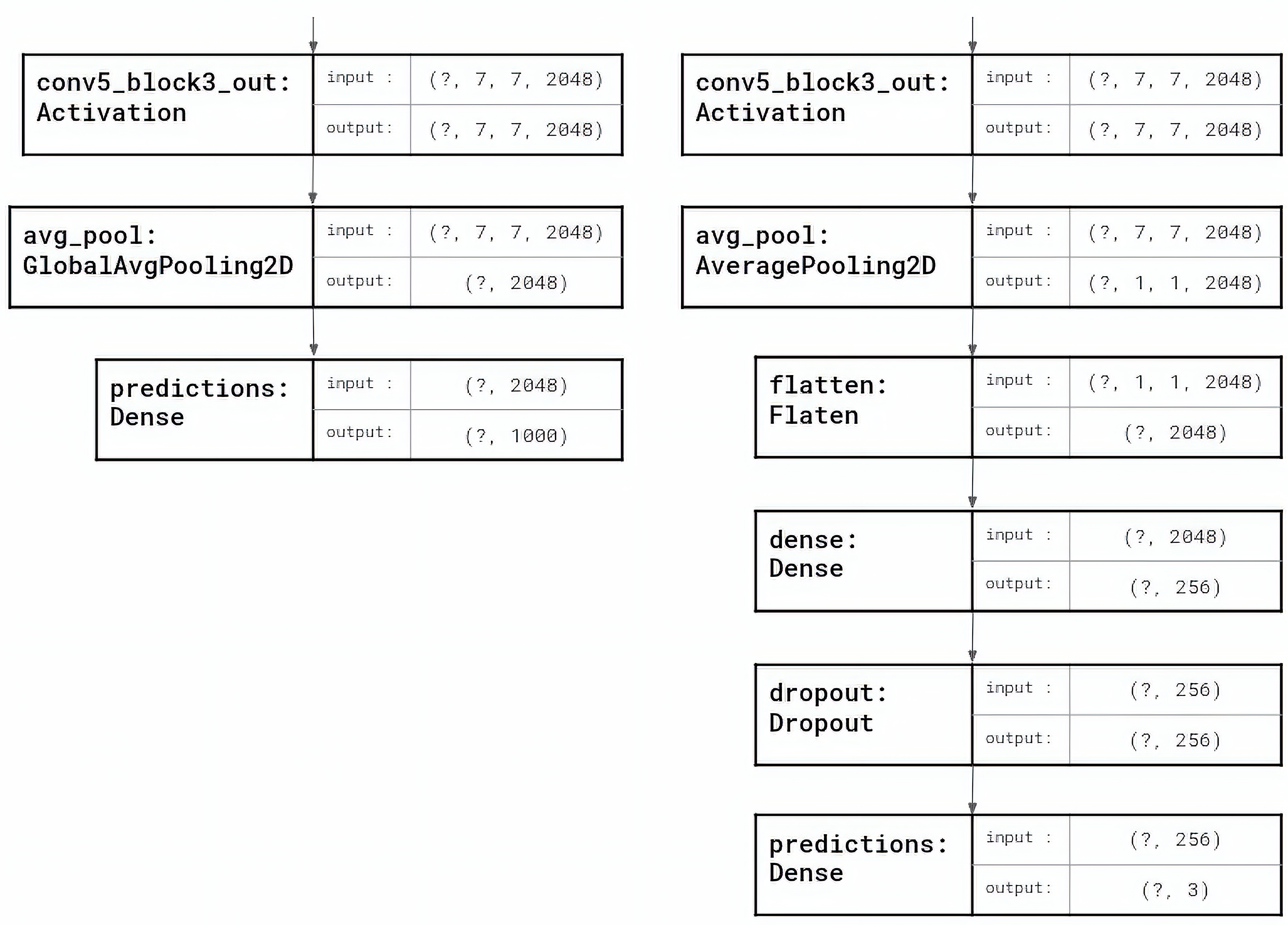}}
			\caption{The last three layers of ResNet50 in their original version are shown on the left. The modifications and layers that were added for the fine tuning on the right.}
		\label{fig3}
	\end{figure}

	To carry out the fine tuning of ResNet50 with our dataset, the following preprocessing was made to the images: resize to 224 x 224 pixels, subtract the mean RGB value of the ImageNet dataset to our images. Comparisons in experiment \#1 in Section \ref{exp1} were done with ResNet50. \\

	\subsection{MobileNet}
	\label{mobilenet_ar}
	
	MobileNetV1 \cite{mobilenetv1} and MobileNetV2 \cite{mobilenetv2} are a family of efficient architectures developed specifically for use on mobile devices, which use convolutions according to the depth. With these networks it was experimented if a lighter architecture exhibits good performance. The modifications made to the last layers of MobileNetV1 are shown in Fig.~\ref{fig4}. \\
	
	\begin{figure}[htbp]
		\centerline{\includegraphics[width=\linewidth]{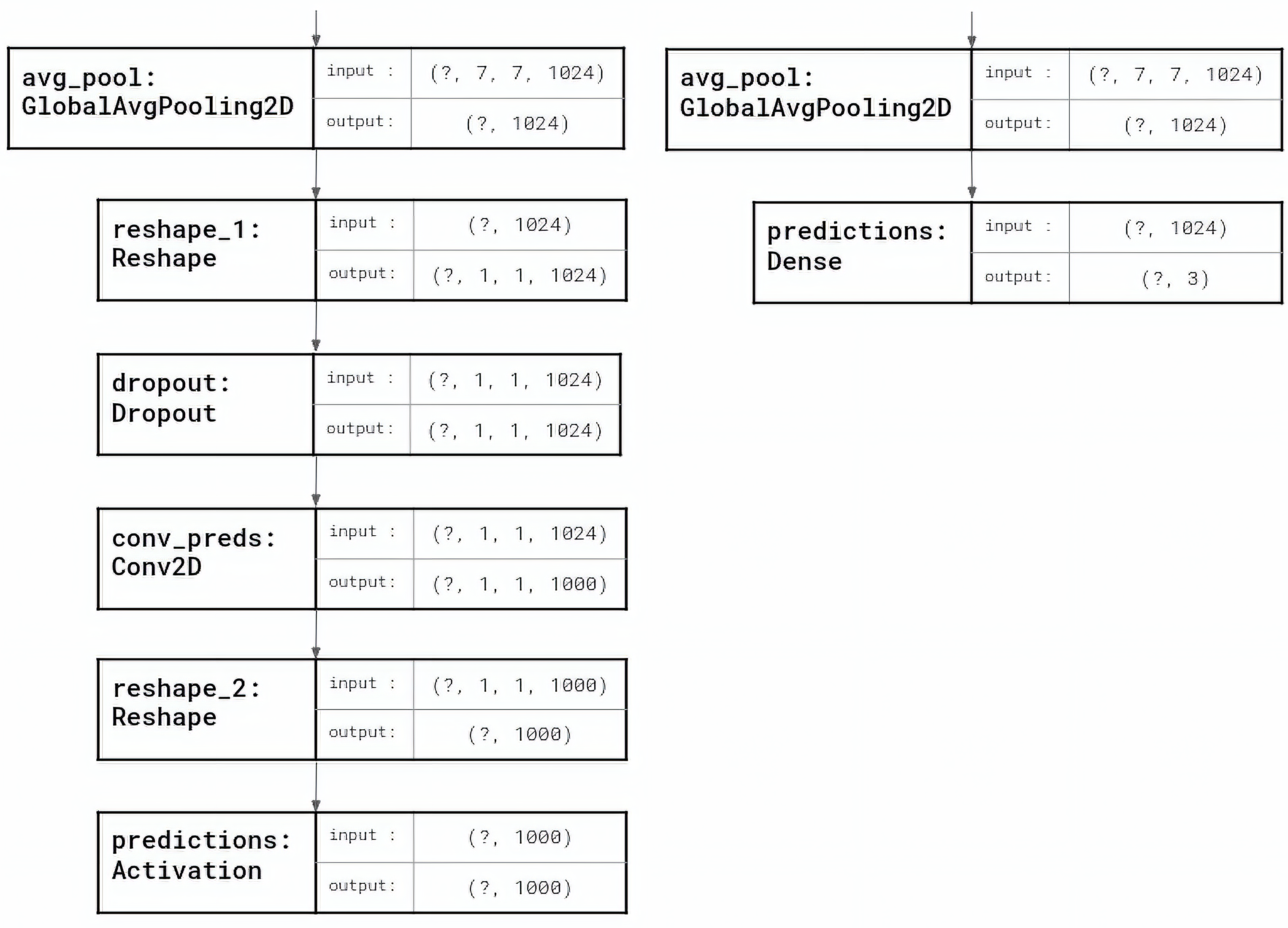}}
		\caption{The last layers of MobileNetV1 in their original version are shown on the left. The modifications and layers that were added for the retraining to the right.}
		\label{fig4}
	\end{figure} 

	And the modifications that were made to the last layers of MobileNetV2 are shown in Fig.~\ref{fig5}. \\
	
	\begin{figure}[htbp]
		\centerline{\includegraphics[width=\linewidth]{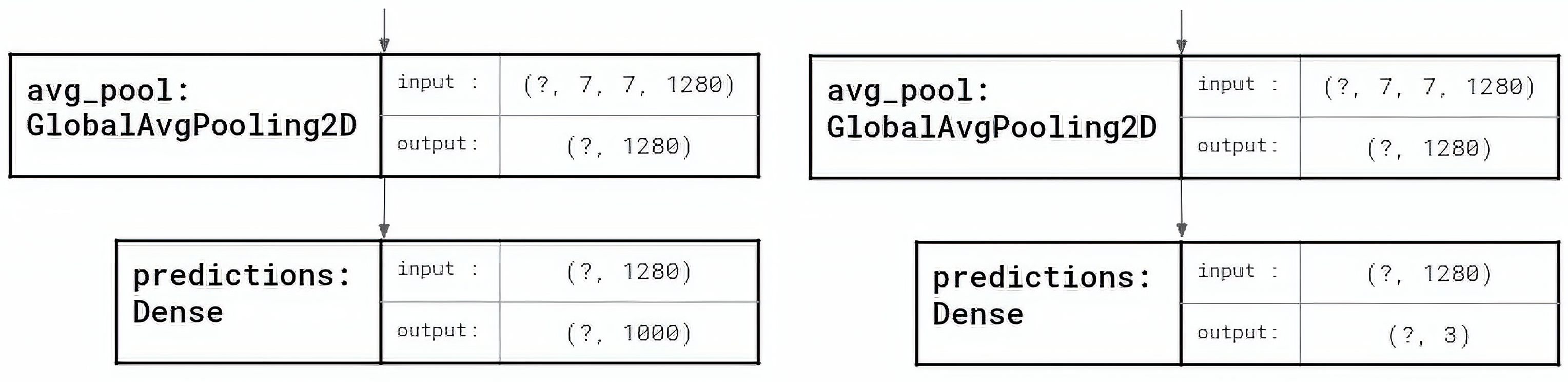}}
		\caption{The last layers of MobileNetV2 in their original version are shown on the left. The modifications and layers that were added for the retraining to the right.}
		\label{fig5}
	\end{figure} 

	MobileNets parameters used are the standard, with $\alpha$\verb| = 1.0| and \verb|input:  224 x 224 pixels|. The processing that the images require is to map the RGB light intensities to the range \verb|(-1 to 1)|. Experiment \#2 in Section \ref{exp2} was performed with this architectures.  \\
	
	In addition to the respective preprocessing, different types of data augmentation were applied to the training set in all experiments. Operations made were: random horizontals turns, random cropping, random zoom and random rotations. \\
	
	The libraries used for the experiments were Tensorflow \cite{tf}, Keras \cite{keras}, Tensorflow Lite for the quantization, OpenCV for preprocessing and scikit-learn for report. The experiments were carried out on the platform Google Colab Pro \cite{googlecolab} with a Tesla P100 Nvidia GPU. \\
	
	In Experiments \#1 and \#2 the various hyperparameters were adjusted until the best result was obtained. Especially the learning rate. The optimizer used is Adam and it used the stepwise decay of the learning rate. For all the training the batch size was 64. \\

	\section{Results and Discussion}
	\label{r_d}
	
	Four experiments were carried out according to the methodology of Section \ref{m_m}. The experiments that obtained the best accuracies in the validation set when adjusting the hyperparameters are presented. Each of them are presented below. \\

	\subsection{\textbf{Experiment \#1}}
	\label{exp1}
	
	ResNet50 was retrained with the images in their original version for 30 epochs. The original layers of the network were frozen. The Adam optimizer was used with initial training rate: \verb|lr = 0.0001| and decay. A 63.5\% accuracy on the validation set was obtained. The loss and accuracy throughout the training is presented in Fig.~ \ref{fig6}. \\
	
	\begin{figure}[htbp]
		\centerline{\includegraphics[width=\linewidth]{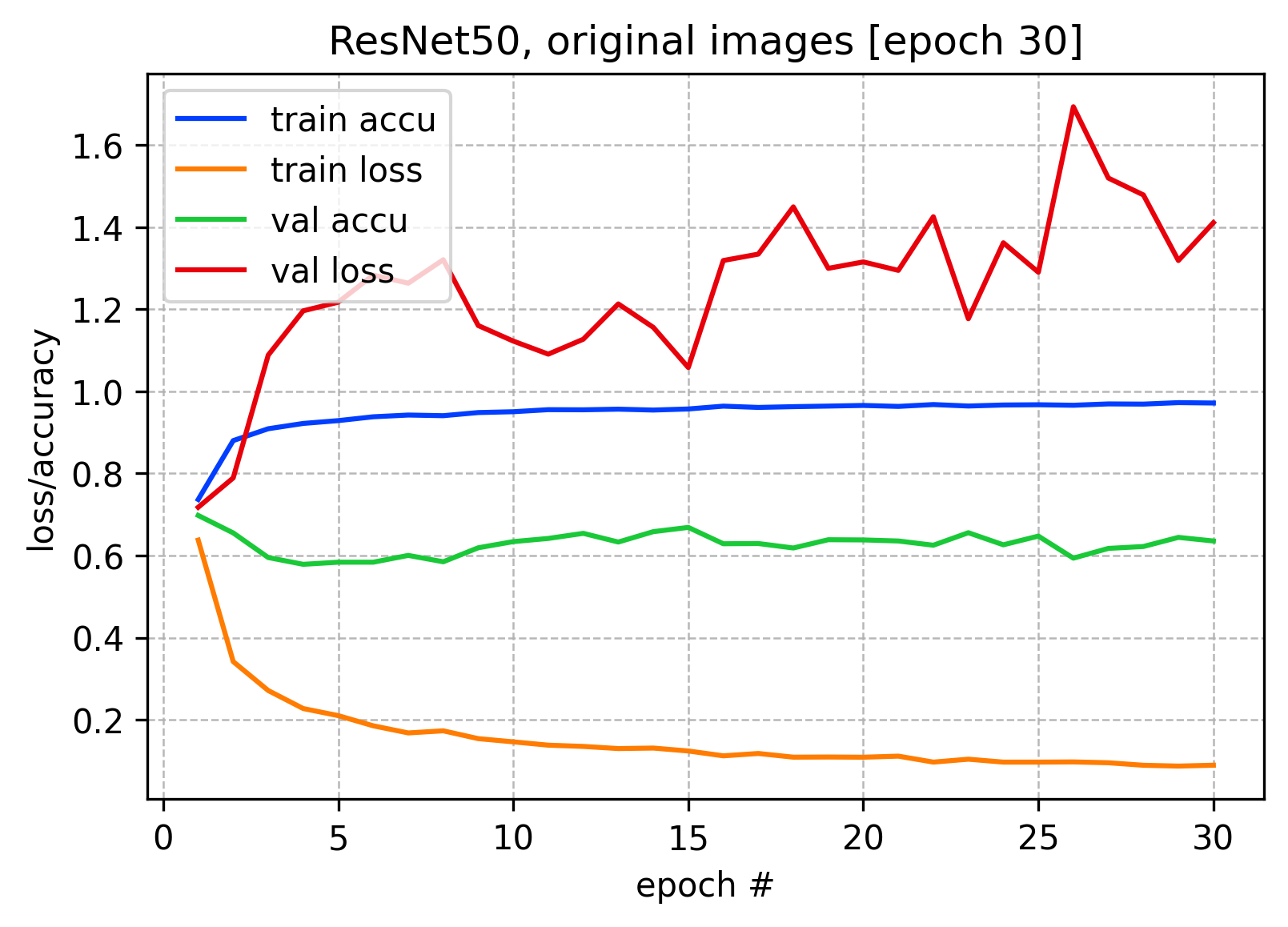}}
		\caption{ResNet50 retraining monitoring, with the ImageNet weights and the Fotini10k dataset in the original version of the images.}
		\label{fig6}
	\end{figure} 

	Then ResNet50 was retrained again, with the same hyperparameters from the first training, but this time using the images in their cropped version. In \verb|epoch 30|, 94.57\% accuracy was obtained in the validation set. Then training was continued, but this time defrosting the last 23 layers and with SDG optimizer and \verb|lr = 0.001| without decay for 50 more epochs. In \verb|epoch 80| the network obtained a 99.21\% accuracy in the validation set. The loss and accuracy throughout the training is presented in Fig.~\ref{fig7}. \\

	\begin{figure}[htbp]
		\centerline{\includegraphics[width=\linewidth]{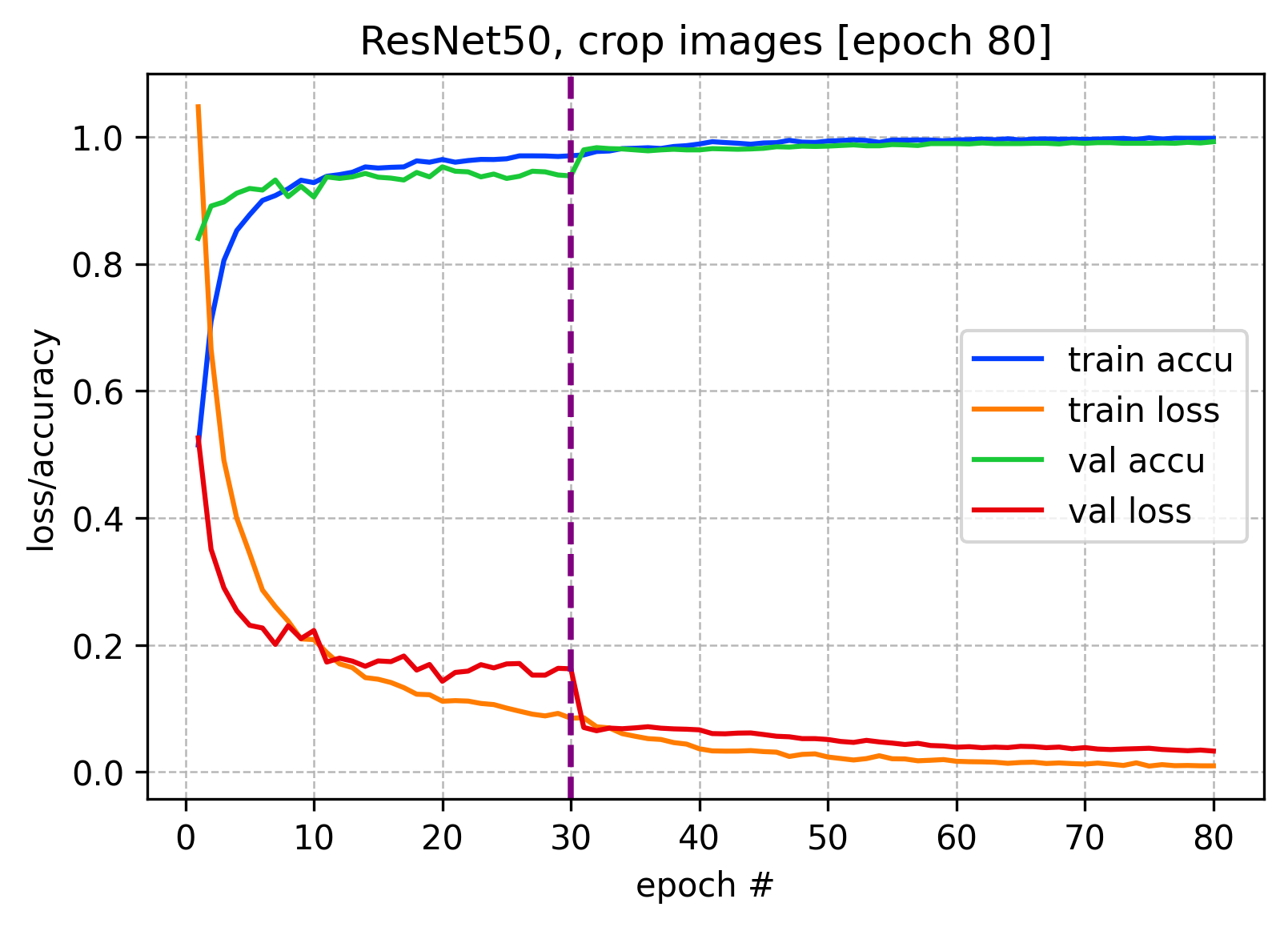}}
		\caption{ResNet50 retraining monitoring, with the ImageNet weights and the Fotini10k dataset in the cropped version of the images. The purple dotted vertical line indicates that from epoch 30 training was done with different hyperparameters.}
		\label{fig7}
	\end{figure} 
	
	\subsection{\textbf{Experiment \#2}}
	\label{exp2}
	
	MobileNetV1-1.0-224 was retrained with the images in their cropped version for 30 epochs. Training of the last 23 layers was enabled, all the previous ones were frozen. The Adam optimizer was used with \verb|lr = 0.0001| and decay. Accuracy of 99.55\% was obtained on the validation set. The loss and accuracy throughout the training is presented in Fig.~\ref{fig8}. \\
	
	\begin{figure}[htbp]
		\centerline{\includegraphics[width=\linewidth]{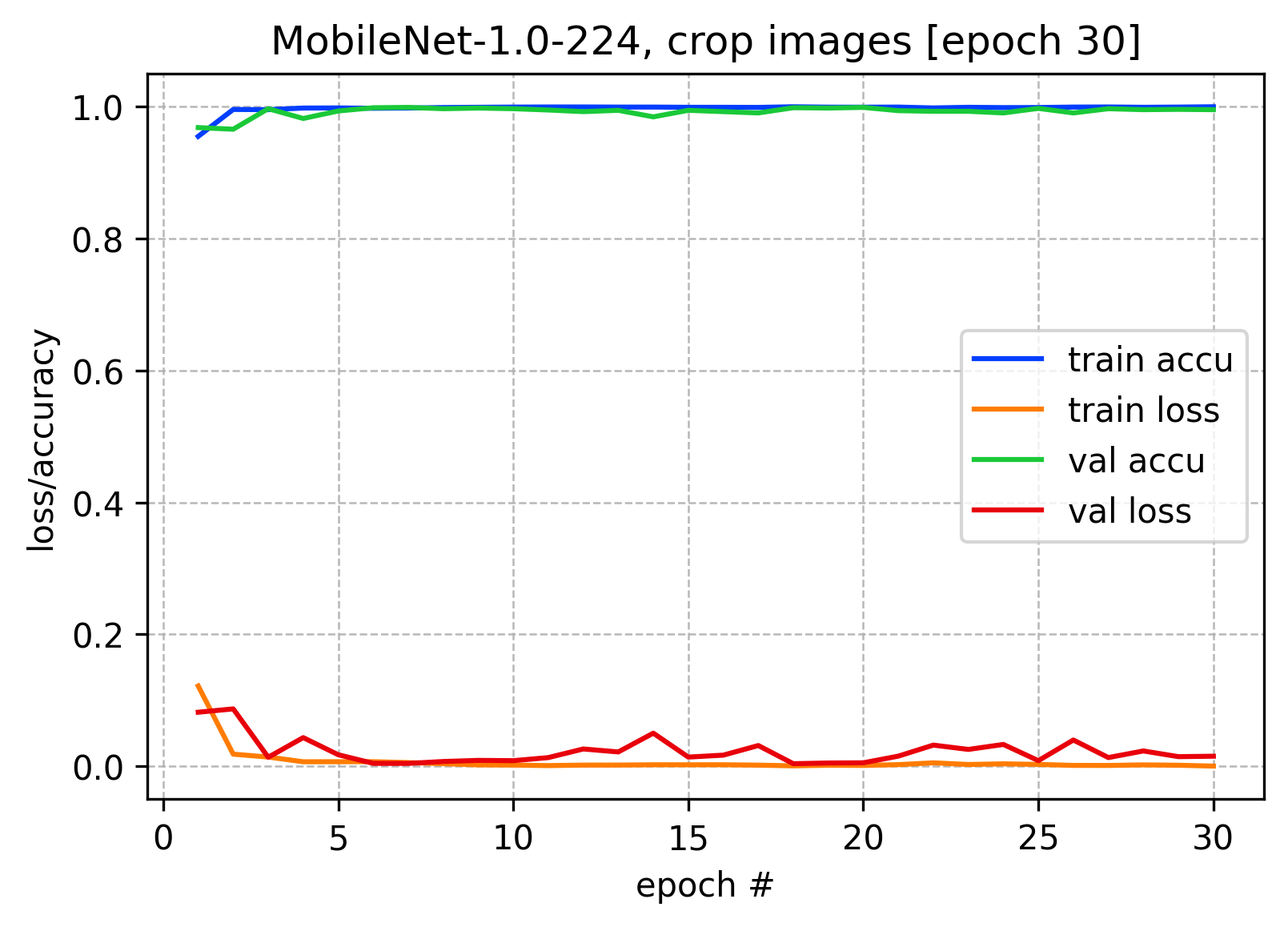}}
		\caption{MobileNetV1 retraining monitoring, with the ImageNet weights and the Fotini10k dataset in the cropped version of the images.}
		\label{fig8}
	\end{figure} 

	MobileNetV2-1.0-224 was retrained with the images in its cropped version for 30 epochs in the same way as MobileNetV1-1.0-224 except that the initial training rate which was set to \verb|lr = 0.00001|. Accuracy of 99.51\% was obtained on the validation set. The loss and accuracy throughout the training is presented in Fig.~\ref{fig9}. \\
	
	\begin{figure}[htbp]
		\centerline{\includegraphics[width=\linewidth]{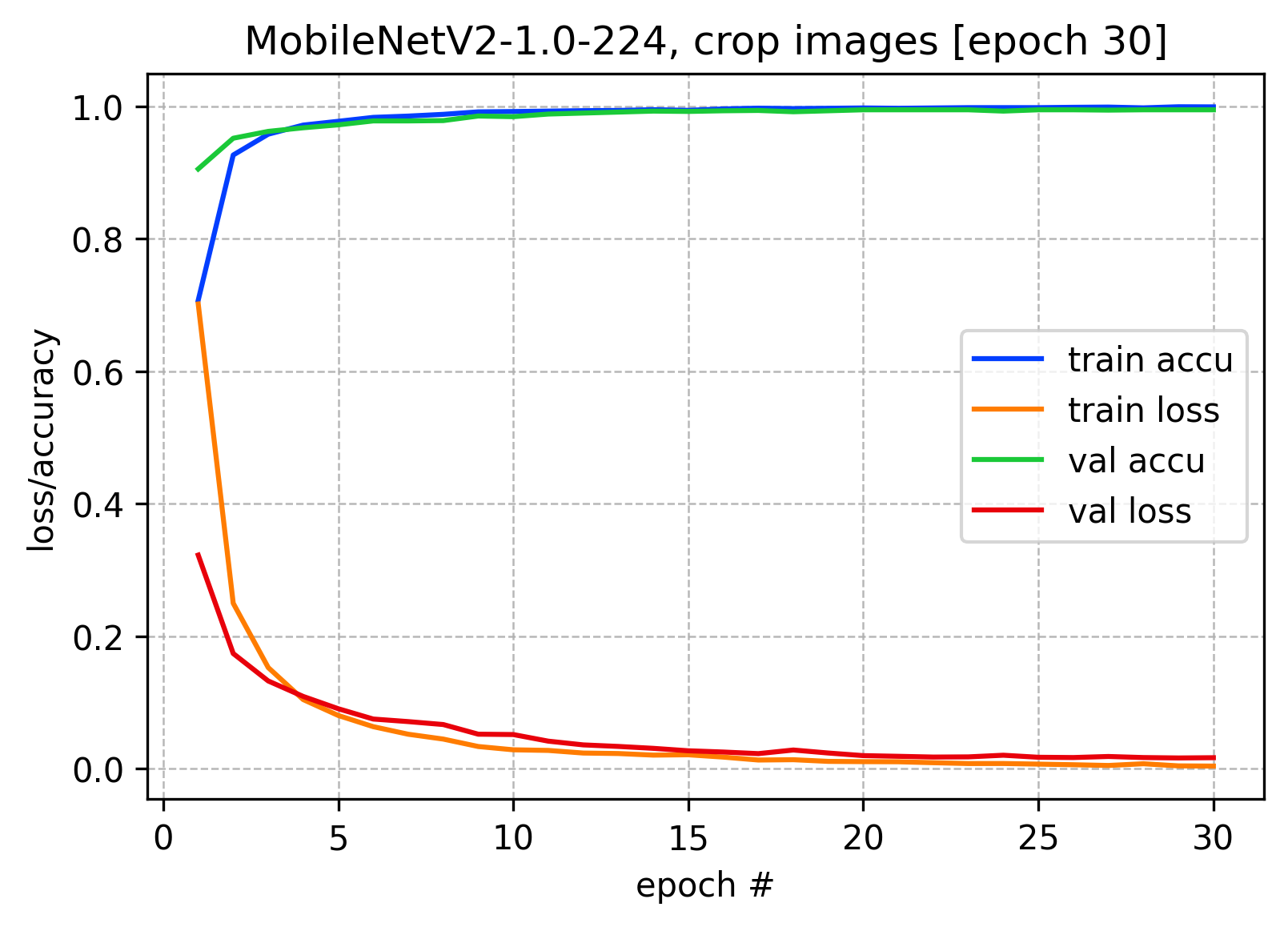}}
		\caption{TMobileNetV2 retraining monitoring, with the Imagenet weights and the Fotini10k dataset in the cropped version of the images.}
		\label{fig9}
	\end{figure} 

 	\pagebreak
	
	\subsection{\textbf{Experiment \#3}}
	\label{exp3}

	After training the three different architectures, each one proceeded to quantize to float16 and ui8 versions. Each architecture, with its variants, was then evaluated in the test set. The results are presented in Table \ref{tab1}. The \verb|8-bit| ResNet50 network is not presented, the reason is presented in Section \ref{three}. \\
	
					\begin{table}[htbp]
						\caption{Top-1 accuracy of the different networks and their variants in the test set (2026 images) of the Fotini10k dataset.}
						\begin{center}
						\renewcommand{\arraystretch}{1.2}
						\begin{tabular}{l  l  l  l  l}
							
							\hline
							\hline
							
							\textbf{} & \textbf{plastic} & \textbf{al cans} & \textbf{paper}  &  \textbf{top-1} \\
							
							\hline
							\hline
							
							\textbf{ResNet50} & 0.99822 & 0.96610  & 0.99570 & 0.98768 \\
							
							\textbf{ResNet50-f16} & 0.99745 & 0.96693 & 0.99506 & 0.98747 \\
							
							\textbf{MobileNetV1} & 1.0000 & 0.99346 & 0.99671 & 0.99602 \\
							
							\textbf{MobileNetV1-f16} & 1.0000 & 0.99032 & 0.99508 & 0.99559 \\
							
							\textbf{MobileNetV1-i8} & 0.99553 & 1.0000 & 0.67595 & 0.89049 \\
							
							\textbf{MobileNetV2} & 0.99169 & 0.99387 & 0.99578 & 0.99333 \\
							
							\textbf{MobileNetV2-f16} & 0.99007 & 0.99346 & 0.99671 & 0.99309 \\
							
							\textbf{MobileNetV2-i8} & 0.99141 & 0.94828 & 0.95374 & 0.96740 \\
							
							\hline
							\hline 
							
						\end{tabular}
						\label{tab1}
						\end{center}	
					\end{table}
				
	Graph in Fig.~\ref{fig10} presents the top-1 accuracy of the different networks in ascending order. \\

	\begin{figure}[htbp]
		\centerline{\includegraphics[width=\linewidth]{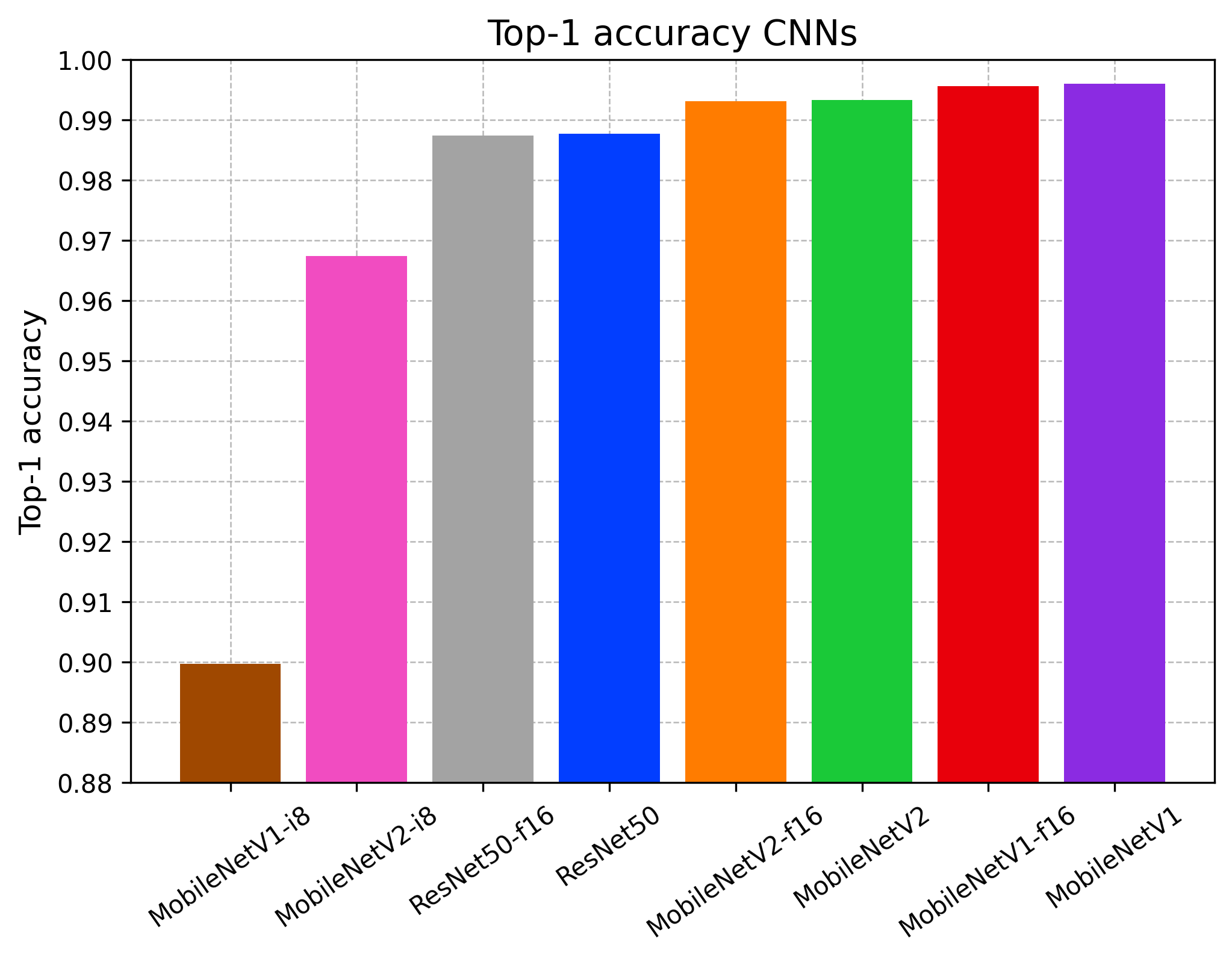}}
		\caption{Top-1 accuracy of the different retrained networks and their quantized variants.}
		\label{fig10}
	\end{figure}

	Graph in Fig.~\ref{fig11} presents the top-1 accuracy of the different architectures and compares them with the number of operations or computational expense (GFlops) that each architecture requires and with their respective memory size. \\	

	\begin{figure}[htbp]
		\centerline{\includegraphics[width=\linewidth]{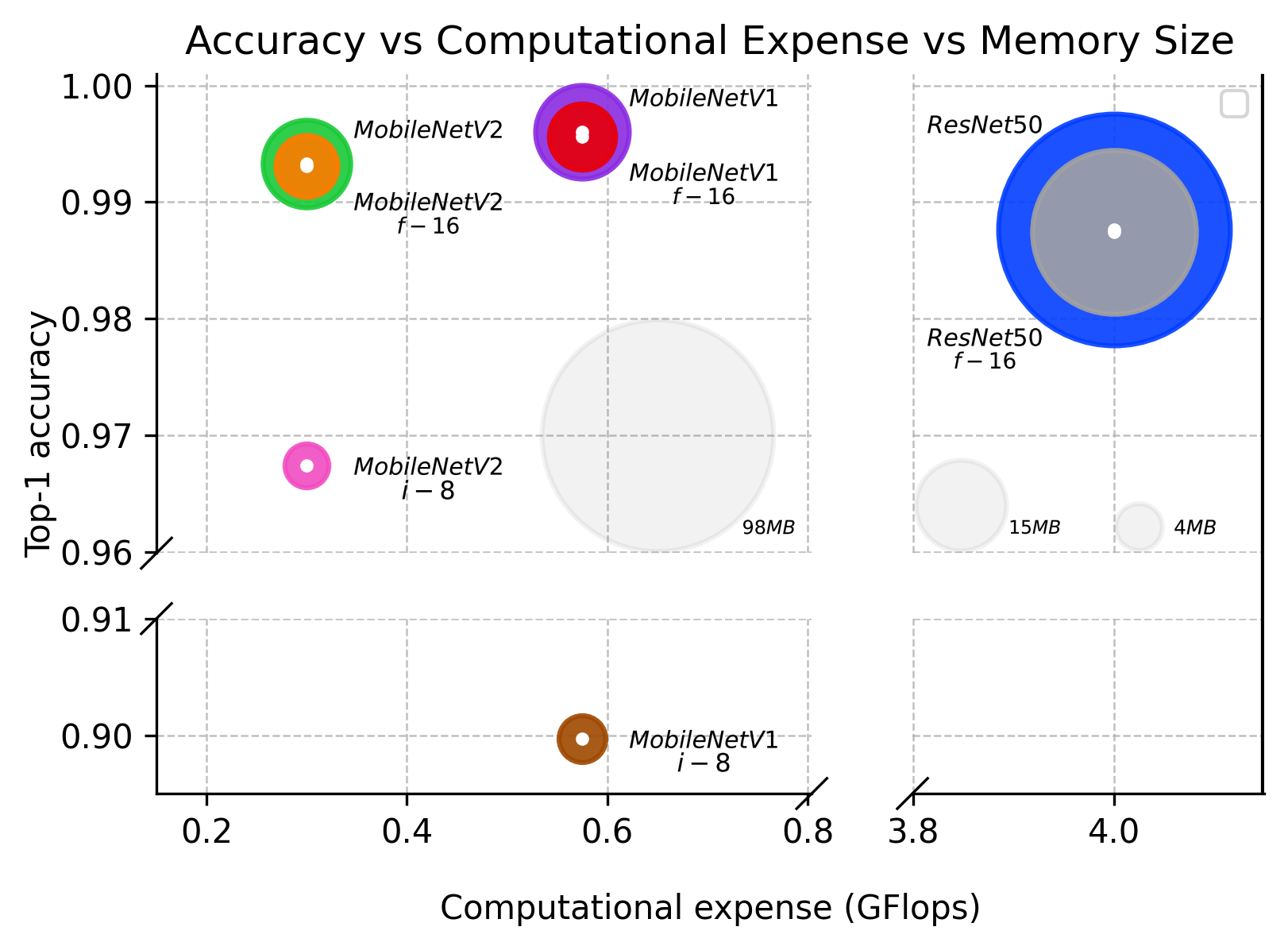}}
		\caption{Top-1 accuracy vs operations to classify an image of the different retrained networks and their variants. The area of the circle represents the memory size of the network. Transparent circles are a memory size reference.}
		\label{fig11}
	\end{figure}	

	\subsection{\textbf{Experiment \#4}}
	\label{exp4}
				
	The last experiment consisted of comparing the inference times (IT) in one image (\verb|batch size = 1|) of each network in some available devices. The measurements shown are the average IT of processing 1000 forward passes the batch size mentioned previously. Not all networks, especially quantized ones, can be used in all devices. An i7-7700HQ CPU, Nvidia GeForce GTX 1050 GPU and Raspberry Pi 4 B 4GB were used for the experiments. The ITs for MobileNetV1-1.0-224 and MobileNetV2-1.0-V2 (quantized to \verb|8-bit|) were also examined on a Google Coral Dev Board. The results are presented in Table \ref{tab2}. Empty cells mean that the network is not supported for that given hardware. A visual comparison is presented in graph on Fig.~\ref{fig12}. It must be taken into account that the scale of the x-axis is logarithmic. \\

					\begin{table}[htbp]
						\caption{Inference time in milliseconds (ms) of Raspberry Pi 4, i7- 7700HQ CPU, Nvidia GTX 1050 GPU and Google Coral Dev board; in the different trained networks and their variants. Empty cells mean that the hardware does not support such a network.}
						\begin{center}
							\renewcommand{\arraystretch}{1.2}
							\begin{tabular}{l  l  l  l  l}
								
								\hline
								\hline
								
								\multicolumn{5}{c}{\textbf{Inference time} (\textit{ms})} \\
							
								\hline
					
								 & RPI 4 & i7-7700 & G1050  &  Coral \\
								
								\hline
								
								\textbf{ResNet50} & 1,244.17 & 121.26 & 84.10 & \\
								
								\textbf{ResNet50-f16} & 952.96 & 76.18 & 63.88 &  \\
								
								\textbf{MobileNetV1} & 352.42 & 45.76 & 42.05 & \\
								
								\textbf{MobileNetV1-f16} & 194.49 & 16.15 & 16.00 & \\
								
								\textbf{MobileNetV1-i8} & & & & 2.40 \\
								
								\textbf{MobileNetV2} & 340.63 & 53.79 & 47.83 & \\
								
								\textbf{MobileNetV2-f16} & 153.41 & 15.68 & 15.59 & \\
								
								\textbf{MobileNetV2-i8} & & & & 2.60 \\
								
								\hline
								\hline 
								
							\end{tabular}
							\label{tab2}
						\end{center}	
					\end{table}

	\pagebreak

	\begin{figure}[htbp]
		\centerline{\includegraphics[width=\linewidth]{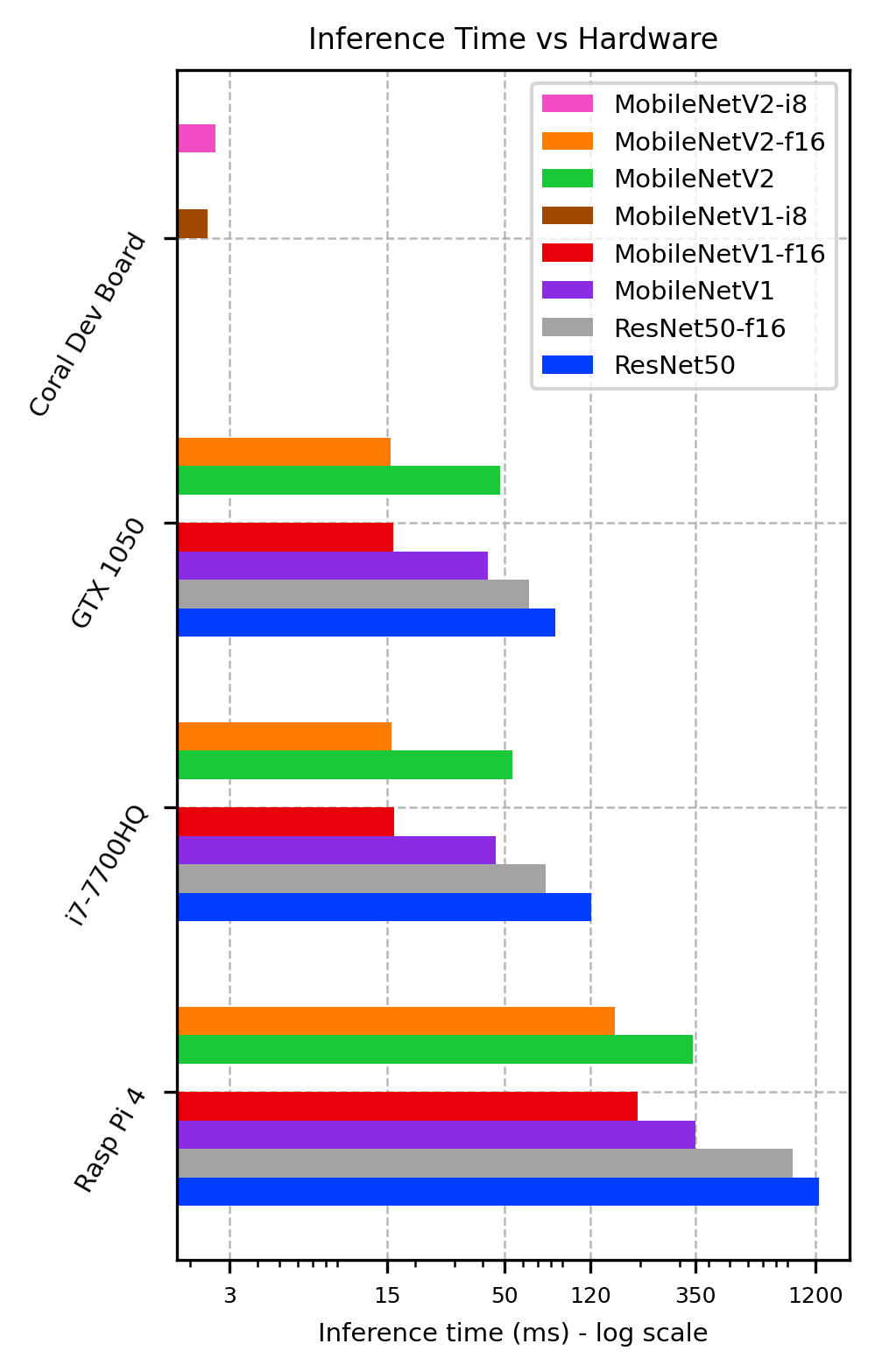}}
		\caption{Comparison of inference times, in milliseconds, of the different hardware proposed in the trained networks and their variants. The x-axis scale is logarithmic.}
		\label{fig12}
	\end{figure}

	\subsection{Discussion}
	
	The results of the experiments and how they can be used in certain applications are discussed below given the advantages and disadvantages of different architectures. \\

	\subsubsection{\textbf{Experiment \#1}}
	\label{three}
	
	Training of ResNet50, with the images in their original version, Graph in Fig.~\ref{fig6}, the network was learning to infer correctly, but only in the training set. When evaluating in the validation set the network could not infer correctly throughout all 30 epochs. This is a phenomenon known as overfitting, which is that the network is not capable of generalizing on data other than the used in training. It is deduced when evaluating the difference of the training curves with the validation one. One possible cause of this is that the images in their original form are very similar despite having different objects. The object is just a small region of the image. The network may be trying to learn to classify the background based on the different spots of light, such as those seen in Fig.~\ref{fig1}, and not on the object itself. The poor performance of the network is verified with the accuracy in the validation set at the end of the training, 63.5\%. \\
	
	In ResNet50 training with the images in its cropped version, Graph on Fig.~\ref{fig7}, the training and validation curves coincide to a great extent. From epoch 20 they start to diverge a bit. At the end of epoch 30, the validation set accuracy is 94.57\%, much better than the best result of 87.8\% from Gyawali et al. \cite{nepal} with ResNet18. With this result, the proposed hypothesis is verified. IC, despite being easier to implement, will not provide the performance of OD, which is much more accurate for the automatic waste classification. The other proposed hypothesis of focusing on classifying objects whose category depends on their visual appearance and not on their composition, is also verified. Better results can be obtained with a dataset with the categories proposed in this work. \\
	
	Using ResNet50 instead of ResNet18 can bring better performance if the dataset used in training is large, like the one in this work. ResNet50 has more layers and therefore can learn more features. Despite satisfactory performance, ResNet50 layers are capable of learning unique characteristics of images, but at the expense of extended training. The first 30 epochs done in this experiment are considered as the warm-up of the new layers that were added in Fig.~\ref{fig3}, since the parameters of these were started randomly. For this reason, ResNet50 was trained for 50 more epochs. In addition, the last 23 layers were defrozed, this allows the network to learn specific characteristics of the dataset from this work. The optimizer was changed from Adam to SDG as Adam is used when looking for fast convergence and SDG when you want the network to learn significant features, regardless of time. As training begins, the training and validation curves immediately begin to converge and match again, a sign that the defrozed layers are learning successfully. At the end of epoch 80, 99.21\% top-1 accuracy was obtained in the validation test. \\
	
	It is important to mention that the reason why 20\% of images were used for validation and later 20\% more for testing, is because the networks will always converge in the training dataset. But this does not mean that they are fulfilling the main objective which is to generalize on unseen data. This is verified by observing the training curves in Graph on Fig.~\ref{fig6} and Graph on Fig.~\ref{fig7}. The two converge, but the performance in the validation set is very different for the two training sessions. By using 20\% of the images on validation, then 20\% more in test in Experiment 3 in Section \ref{exp3}, we are confident of our results; and when used in a real application, the performance will be expected to be effectively the same. \\

	\subsubsection{\textbf{Experiment \#2}}
	
	The performance in the previous training is excellent, but it has the disadvantage that ResNet50 is a network that demands a lot of computational resources and memory space. This makes it not an ideal network to be used on mobile devices. For this reason, experiments were done with the Fotini10k dataset and the MobileNets family of networks, which offer reduced computational expenses. \\
	
	The first 23 layers of MobileNetV1 and MobileNetV2 were defrozed from the beginning, since their layers do not need as extensive training as ResNet50. It is not even necessary to add extra layers of generalization as seen in the modifications to the MobileNets networks in Fig.\ref{fig4} and \ref{fig5} vs ResNet50 in Fig.~\ref{fig3}. Looking at Fig.~\ref{fig6} and \ref{fig7}, the curves coincide to a great extent. This means that they are not overfitting. At the end of epoch 30,  99.55\% accuracy is obtained on MobileNetV1 and 99.51\% on MobileNetV2. With these results it is verified that the performance of the Fotini10k dataset is maintained even with light architectures. \\
	
	The same performance was obtained with the MobileNets, with much less work (30 epoch vs 80 epoch), than with ResNet50 because indeed the Fotini10k dataset is very robust. But, it also plays an important role that is only being discerned between three categories. A lightweight network can learn the necessary discriminative characteristics with few epochs. A more robust network like ResNet50 requires more parameters to be adjusted. However, in the case of adding more categories or categories that are more difficult to distinguish from each other, PET plastic 1 vs PET plastic 2 for example, ResNet50 can maintain the accuracy. MobileNets could become saturated as its layers could not capture as much information. This is a hypothetical case to consider in future research and serves as a guide to know which network to use according to the type of implementation looked for.
	MobileNetV1 and MobileNetV2 show identical results, despite MobileNetV2 improvements \cite{mobilenetv2} over MobileNetV1. This is expected as the Fotini10k dataset gives enough examples to achieve really good performance and it is difficult to see significant differences between one network and another, especially in classification performance. Despite this, some differences become apparent in upcoming discussions when evaluating its performance in its quantized versions and memory space. \\

	\subsubsection{\textbf{Experiment \#3}}
	
	This experiment consists of the quantization of the three architectures already trained to float16 and ui8. The specific technique used is post-training quantization \cite{quantf}. This means that the already trained network is taken in standard form, a \verb|.hdf5| file, and the conversions are carried out through the Tensorflow Lite library. The converted network has extension \verb|.tflite|. \\
	
	Post-training quantization differs from pre-training quantization, since in the second, the network training is performed with the parameters already quantized. The pre-training method provides better results for \verb|int-8| quantizations and is even strictly necessary to implement such quantized networks in \verb|int-8| TPU architecture and microcontrollers. Even so, all the post-training quantizations were carried out on the trained networks, because they serve as the basis for knowing the precision that can be expected and even improve in \verb|int-8| quantizations. Conversions to float16 do not present these details. Although it is important to emphasize that to effectively perform inference with float16 parameters, it is necessary to use a GPU (graphics processing unit), since they support these data types in their hardware. On a CPU the parameters will be converted to float32 anyway. \\
	
	It is important to mention that when converting the network to \verb|.tflite|, optimizations are done such that IT is improved, including when the network weights are converted back to float32. But this implies that the precision may possibly decrease. The ITs are objected in the discussions of experiment \#4 in Section \ref{exp4}. \\
	
	Once all the networks had been quantized, they were evaluated in their standard form and quantized versions. The metric for this evaluation was top-1 accuracy. The images that were evaluated are the test dataset, 20\% (2026 images) of the total images. These images were never seen by the network in training. This ensures a true performance metric that can be expected in real applications. The results are presented in Table \ref{tab1}. \\
	Before discussing the results, ResNet50 supported quantization to \verb|int-8| but its results are not present in the graphs as they are null. Lesser precision than could be obtained if it is chosen at random. This verifies that ResNet50 is not well supported for mobile devices; at least not with the software and hardware easily available now. \\
	
	Fig.~\ref{fig10} shows in ascending order the different top-1 accuracies of the networks. Nets in their standard and float16 form do not have top-1 accuracy lower than 98.7\%, effectively 99\%. In the comparison of the networks in their standard form vs float16, the precision loss is insignificant, less than 0.001\% in any of the three pairs. When looking at the precision of the networks quantized to \verb|int-8|, MobileNetV1-i8 has an accuracy of 89.0\% and MobileNetV2-i8 96.7\%. The accuracy loss in MobileNetV2-i8 is 2\%. The accuracy drop of MobileNetV1-i8 is 10\%, when inquiring into the accuracies by category, a noticeable imbalance is noted in the category of \verb|'paper and cardboard'| with accuracy of 67.6\% vs 99.6\% and 100\% in the other two categories. \\
	
	These results prove that, in terms of accuracy, the networks developed are robust in light architectures and even in architectures quantized to float16 since they all maintain top-1 accuracies of 99\%. It is important to mention that in networks quantized to \verb|int-8|, the unbalance by class of MobileNetV1-i8 is unfavorable and therefore in applications that require \verb|int-8| for TPU microcontrollers it is favorable to use MobileNetV2-i8. The reasons why MobileNetV2 quantized to \verb|int-8| maintains its accuracy over MobileNetV1 lies in the advantages introduced in \cite{mobilenetv2} that make it possible to maintain precision in these conditions, even occupying only 4 MB in memory and reducing 275 MFlops with respect to MobileNetV1. \\
	
	Fig.~\ref{fig11} illustrates a comparison of the top-1 accuracy of networks with respect to the number of operations they need to perform a classification with \verb|batch size =  1|. The size that the network occupies in memory is also compared. The first observation is the large memory size that ResNet50 occupies, 98MB. MobileNets networks occupy about 15 MB of memory in their standard form. This makes them very favorable for mobile applications. When observing the computational expense, due to the numerous layers of ResNet50, it requires 4 GFlops to process an image, to obtain the same results, the MobileNets require \verb|10x| less operations. This proves the superiority of MobileNets for mobile applications and even standard applications, at least with the Fotini10k dataset. \\
		
	The quantization effects in the different networks, according to Fig.~\ref{fig11}, are a significant reduction in memory, \verb|2x|, maintaining precision of 99\% for quantizations to float16. Quantizing to \verb|int-8| reduces the size by \verb|4x| the original size. MobileNetV1 and MobileNetV2 occupy 4MB and 3MB respectively. For MobileNetV2 this reduction only entails a cost of only 2\% decrease in accuracy. \\
	
	It is observed that when quantizing the number of operations per network is maintained, this is because the architecture or layers of the network are the same. However, operations on float16 and \verb|int-8| take less resources. This is reflected in the IT performance discussed below. \\
	
	\subsubsection{\textbf{Experiment \#4}}
	
	The IT for each network with \verb|batch size =  1|, is an important metric since in most computer vision applications they process the frames of a video input, one by one. How fast a frame can be processed determines, if real-time inference is possible, most likely what is needed for automatic waste sorting. This speed is measured in frames per second fps (the inverse of IT). The amount of fps needed depends specifically on the application. IT can be called the processing delay time, taking into account that the sampling time is negligible with respect to the IT. It is important to mention that depending on the application, the forward pass of the image in the CNN is not the only operation to be carried out, this is true for OD, the application that is being sought. That is why it is important to obtain the lowest possible ITs, taking into account the possible decreases in accuracy. \\
	
	IT depends on the processing unit being used. In experiment \#4 in Section {ref{exp4}, the mentioned hardware was used since they were the available ones. The Raspberry Pi 4 represents a possible candidate for a mobile application. This experiment served primarily as a frame of reference to determine the IT capabilities of the trained networks. The IT may fluctuate a bit, for this reason the results obtained are an average of 1000 inferences. Results are shown in Table \ref{tab2}. \\
		
	Fig.~\ref{fig12} shows a comparison of the results. The logarithmic scale was used, as the ITs on the Raspberry Pi are very high relative to the others. ResNet50 requires 1240 ms to infer. Even ResNet50-f16 requires 953 ms. This in turn checks the limitations of ResNet50 for use in mobile applications. The ITs of the MobileNets on the Raspberry Pi 4 are still high, but variants quantized to float16 present ITs of 194 ms for MobileNetV1-f16 and 153 ms for MobileNetV2-i8. This is around 6 to 7 fps, which can be useful for simple applications. Using an i7-7700HQ CPU, it was possible to reduce IT. ResNet50-f16 has an IT of 76.18 ms. MobileNets feature 48 ms IT on average in their standard variant. Variants quantized to float16 present TI of 16 ms. This is around 60 fps, a very useful performance, especially considering that these networks have 99\% top-1 accuracy. When comparing the results of the i7-7700HQ CPU with those of the Nvidia GTX 1050 GPU, it is observed that the IT does not improve significantly. This is because one image is being processed at a time and thus the GPU cannot use its main advantage, which is parallel computing. Also, this GPU is very low-end. \\
	
	In general, it can be seen that MobileNets actually present much lower IT than ResNet50. Also, when using the network quantized to float16 the ITs decrease about \verb|2x|. This proves the usefulness of network quantizations in this particular application, because the decrease in accuracy is virtually null with networks trained on the Fotini10k dataset. \\
	
	Finally, the performance that can be obtained when using networks quantized to \verb|int-8| was discussed. The Google Coral Dev Board device has a TPU that exclusively uses networks with pre-training quantization, in Fig.~\ref{fig12} it is observed that this device has an IT of 2.60 ms in MobileNetV2 \cite{coralll}. Despite this result not being tested in the specifically trained network, due to availability issues, the same IT would be had with the MobileNetV2-i8 network of this investigation since it is exactly the same architecture and therefore the same operations, only different weight values. This represents a speed of 350+ fps and our MobileNetV2-i8 network that has a top-1 accuracy of 97\%. For future research and applications it is possible to evaluate other hardware, including the Jetson Nano Dev Kit, the Coral TPU accelerator that can be used in conjunction with the Raspberry Pi 4. \\
	
	It is important to mention that MobileNets networks can even be made more efficient and smaller by reducing the input resolution and parameter $\alpha$ \cite{mobilenetv1}, in this way they can be used even in microcontrollers with Cortex-M4 or M7 chips, for example. The cost of these adjustments is the reduction in accuracy Networks trained in the Fotini10k dataset show great tolerance on accuracy drop to these adjustments. Future research will evaluate these possibilities and their limitations. \\
	
	The latest experiments serve as a guide to which network to use for each task, and performance on each hardware. It should be noted that the goal is to develop OD. These results can be used for applications that only require IC. In the future research OD will be developed and evaluated with all networks, but especially MobileNetV2 that gave the best results and it is very common to use it in conjunction with the SSD technique to perform OD. On average the training took 86 to 100 seconds per epoch. The dataset and networks will be published soon at: \url{https://github.com/jaimix4}.

	\section{Conclusions}
	\label{concl}
	
	In this work it has been concluded that: \\
	
	\begin{itemize}
		
	\item Although image classification is easier to implement, object detection provides a more robust solution for automatic waste classification.
	\item When using CNN in images to classify waste, better results are obtained when classifying categories that depend on the visual appearance and not on the composition of the debris.
	\item The Fotini10k dataset, despite being presented as a tool at the beginning of this research, becomes another result when its performance is validated in the different networks previously trained in ImageNet. By taking the images in a visually neutral enclosure with the robot built, a large number of annotated images were obtained. These are representative of real situations, for the categories: \verb|'aluminum cans', 'plastic bottles'| and \verb|'paper and cardboard'|. The performance of our CNNs is a state of the art result for the task of waste image classification.
	\item The retraining provided state-of-the-art results across all three architectures (ResNet50, MobileNetV1, and MobileNetV2 with Imagenet weights) by achieving 99\% accuracy and exceeding Gyawali et al's previous score of 88\% \cite{nepal}.
	\item ResNet50 requires more specialized training compared to MobileNets to achieve the same results. This is especially true for a few category models.
	\item Training with the Fotini10k dataset are very tolerant of quantization operations, decreasing only 0.001\% when applying quantization to float16 in different architectures. The MobileNetV2 network when quantized to \verb|8-bit| weights, only presented a 2\% decrease in precision (99\% original). This makes the Fotini10K dataset ideal for use in mobile applications.
	\item The networks presented in this work present useful inference times even on Raspberry Pi 4 as it is possible to obtain 6 to 7 fps with MobileNets quantized in a .tflite.
	\item For applications where computational power is not a limitation, the MobileNets networks presented in this research obtain 60+ fps in standard CPUs when quantized in .tflite (float16) and top-1 precision of 99\%.
	\item MobileNetV2 quantized in \verb|int-8| can be implemented with inference speeds of 350+ fps and 97\% accuracy. A state-of-the-art result for automatic waste sorting with CNNs.
	\item The results of this research are extremely useful for image classification in waste and serve as the basis for developing object detection. Future research will explore the SSD + MobileNetV2 combination with the Fotini10k dataset.

	\end{itemize}

	\section*{Acknowledgment}
	
	We thank Professor Matteo Matteucci and the AIRLAB laboratory of the Politecnico di Milano for their help in developing the Fotini10k dataset on their premises at Campus Leonardo, Milan, Italy. We are also grateful to the different professors of the Faculty of Electrical Engineering of the Regional Center of Chiriqui, Technological University of Panama for their help in developing the idea through academic research and development projects. \\


	\bibliographystyle{IEEEtran}
	\bibliography{references}

\begin{thebibliography}{10}
\providecommand{\url}[1]{#1}
\csname url@samestyle\endcsname
\providecommand{\newblock}{\relax}
\providecommand{\bibinfo}[2]{#2}
\providecommand{\BIBentrySTDinterwordspacing}{\spaceskip=0pt\relax}
\providecommand{\BIBentryALTinterwordstretchfactor}{4}
\providecommand{\BIBentryALTinterwordspacing}{\spaceskip=\fontdimen2\font plus
\BIBentryALTinterwordstretchfactor\fontdimen3\font minus
  \fontdimen4\font\relax}
\providecommand{\BIBforeignlanguage}[2]{{%
\expandafter\ifx\csname l@#1\endcsname\relax
\typeout{** WARNING: IEEEtran.bst: No hyphenation pattern has been}%
\typeout{** loaded for the language `#1'. Using the pattern for}%
\typeout{** the default language instead.}%
\else
\language=\csname l@#1\endcsname
\fi
#2}}
\providecommand{\BIBdecl}{\relax}
\BIBdecl

\bibitem{kaza2018waste}
S.~Kaza, L.~Yao, P.~Bhada-Tata, and F.~Van~Woerden, \emph{What a waste 2.0: a
  global snapshot of solid waste management to 2050}.\hskip 1em plus 0.5em
  minus 0.4em\relax The World Bank, 2018.

\bibitem{ferronato2019waste}
N.~Ferronato and V.~Torretta, ``Waste mismanagement in developing countries: A
  review of global issues,'' \emph{International journal of environmental
  research and public health}, vol.~16, no.~6, p. 1060, 2019.

\bibitem{alexnet}
A.~Krizhevsky, I.~Sutskever, and G.~E. Hinton, ``Imagenet classification with
  deep convolutional neural networks,'' in \emph{Advances in neural information
  processing systems}, 2012, pp. 1097--1105.

\bibitem{imagenet}
O.~Russakovsky, J.~Deng, H.~Su, J.~Krause, S.~Satheesh, S.~Ma, Z.~Huang,
  A.~Karpathy, A.~Khosla, M.~Bernstein \emph{et~al.}, ``Imagenet large scale
  visual recognition challenge,'' \emph{International journal of computer
  vision}, vol. 115, no.~3, pp. 211--252, 2015.

\bibitem{vgg}
K.~Simonyan and A.~Zisserman, ``Very deep convolutional networks for
  large-scale image recognition,'' \emph{arXiv preprint arXiv:1409.1556}, 2014.

\bibitem{googlenet}
C.~Szegedy, W.~Liu, Y.~Jia, P.~Sermanet, S.~Reed, D.~Anguelov, D.~Erhan,
  V.~Vanhoucke, and A.~Rabinovich, ``Going deeper with convolutions,'' in
  \emph{Proceedings of the IEEE conference on computer vision and pattern
  recognition}, 2015, pp. 1--9.

\bibitem{resnet}
K.~He, X.~Zhang, S.~Ren, and J.~Sun, ``Identity mappings in deep residual
  networks,'' in \emph{European conference on computer vision}.\hskip 1em plus
  0.5em minus 0.4em\relax Springer, 2016, pp. 630--645.

\bibitem{stanford}
\BIBentryALTinterwordspacing
O.~Awe, R.~Mengistu, and V.~Sreedhar, ``{Final Report : Smart Trash Net : Waste
  Localization and Classification},'' Tech. Rep., 2017. [Online]. Available:
  \url{http://cs229.stanford.edu/proj2017/final-reports/5226723.pdf}
\BIBentrySTDinterwordspacing

\bibitem{nepal}
\BIBentryALTinterwordspacing
D.~Gyawali, A.~Regmi, A.~Shakya, A.~Gautam, and S.~Shrestha, ``{Comparative
  Analysis of Multiple Deep CNN Models for Waste Classification},'' apr 2020.
  [Online]. Available: \url{http://arxiv.org/abs/2004.02168}
\BIBentrySTDinterwordspacing

\bibitem{rosebrock_dl4cv}
A.~Rosebrock, \emph{Deep Learning for Computer Vision with Python},
  3rd~ed.\hskip 1em plus 0.5em minus 0.4em\relax PyImageSearch.com, 2019.

\bibitem{adam}
D.~P. Kingma and J.~Ba, ``Adam: A method for stochastic optimization,''
  \emph{arXiv preprint arXiv:1412.6980}, 2014.

\bibitem{rmsprop}
\BIBentryALTinterwordspacing
G.~Hinton, N.~Srivastava, and K.~Swersky, ``{Neural Networks for Machine
  Learning Lecture 6a Overview of mini-­-batch gradient descent},'' pp. 13,
  87. [Online]. Available:
  \url{https://www.cs.toronto.edu/~tijmen/csc321/slides/lecture_slides_lec6.pdf}
\BIBentrySTDinterwordspacing

\bibitem{coco}
T.-Y. Lin, M.~Maire, S.~Belongie, J.~Hays, P.~Perona, D.~Ramanan,
  P.~Doll{\'a}r, and C.~L. Zitnick, ``Microsoft coco: Common objects in
  context,'' in \emph{European conference on computer vision}.\hskip 1em plus
  0.5em minus 0.4em\relax Springer, 2014, pp. 740--755.

\bibitem{ssd}
W.~Liu, D.~Anguelov, D.~Erhan, C.~Szegedy, S.~Reed, C.-Y. Fu, and A.~C. Berg,
  ``Ssd: Single shot multibox detector,'' in \emph{European conference on
  computer vision}.\hskip 1em plus 0.5em minus 0.4em\relax Springer, 2016, pp.
  21--37.

\bibitem{quan}
B.~Jacob, S.~Kligys, B.~Chen, M.~Zhu, M.~Tang, A.~Howard, H.~Adam, and
  D.~Kalenichenko, ``Quantization and training of neural networks for efficient
  integer-arithmetic-only inference,'' in \emph{Proceedings of the IEEE
  Conference on Computer Vision and Pattern Recognition}, 2018, pp. 2704--2713.

\bibitem{mobilenetv1}
A.~G. Howard, M.~Zhu, B.~Chen, D.~Kalenichenko, W.~Wang, T.~Weyand,
  M.~Andreetto, and H.~Adam, ``Mobilenets: Efficient convolutional neural
  networks for mobile vision applications,'' \emph{arXiv preprint
  arXiv:1704.04861}, 2017.

\bibitem{mobilenetv2}
M.~Sandler, A.~Howard, M.~Zhu, A.~Zhmoginov, and L.-C. Chen, ``Mobilenetv2:
  Inverted residuals and linear bottlenecks,'' in \emph{Proceedings of the IEEE
  conference on computer vision and pattern recognition}, 2018, pp. 4510--4520.

\bibitem{tf}
\BIBentryALTinterwordspacing
M.~Abadi, A.~Agarwal, P.~Barham, E.~Brevdo, Z.~Chen, C.~Citro, G.~S. Corrado,
  A.~Davis, J.~Dean, M.~Devin, S.~Ghemawat, I.~Goodfellow, A.~Harp, G.~Irving,
  M.~Isard, Y.~Jia, R.~Jozefowicz, L.~Kaiser, M.~Kudlur, J.~Levenberg,
  D.~Man\'{e}, R.~Monga, S.~Moore, D.~Murray, C.~Olah, M.~Schuster, J.~Shlens,
  B.~Steiner, I.~Sutskever, K.~Talwar, P.~Tucker, V.~Vanhoucke, V.~Vasudevan,
  F.~Vi\'{e}gas, O.~Vinyals, P.~Warden, M.~Wattenberg, M.~Wicke, Y.~Yu, and
  X.~Zheng, ``{TensorFlow}: Large-scale machine learning on heterogeneous
  systems,'' 2015, software available from tensorflow.org. [Online]. Available:
  \url{https://www.tensorflow.org/}
\BIBentrySTDinterwordspacing

\bibitem{keras}
F.~Chollet \emph{et~al.}, ``Keras,'' \url{https://keras.io}, 2015.

\bibitem{googlecolab}
E.~Bisong, ``Google colaboratory,'' in \emph{Building Machine Learning and Deep
  Learning Models on Google Cloud Platform}.\hskip 1em plus 0.5em minus
  0.4em\relax Springer, 2019, pp. 59--64.

\bibitem{quantf}
\BIBentryALTinterwordspacing
``{Post-training quantization  |  TensorFlow Lite}.'' [Online]. Available:
  \url{https://www.tensorflow.org/lite/performance/post_training_quantization}
\BIBentrySTDinterwordspacing

\bibitem{coralll}
\BIBentryALTinterwordspacing
``{Edge TPU performance benchmarks | Coral}.'' [Online]. Available:
  \url{https://coral.ai/docs/edgetpu/benchmarks/}
\BIBentrySTDinterwordspacing

\end{thebibliography}

\end{document}